\documentclass[sigconf, screen]{acmart}

\AtBeginDocument{%
  }

\usepackage{pifont}

\usepackage{hyperref}
\usepackage{url}
\usepackage{bm}
\usepackage{enumitem}
\usepackage{wrapfig}
\usepackage{booktabs}
\usepackage{multirow}
\usepackage{caption}
\usepackage{graphicx}
\usepackage{subfigure}
\usepackage{wrapfig}
\usepackage{mathtools}
\usepackage{amsmath,amsfonts, amsthm}
\usepackage{tikz}
\usepackage{algorithm}
\usepackage{algpseudocode}
\usepackage{threeparttable}
\usepackage{xspace}
\usepackage{tcolorbox}
\usepackage{subcaption}
\usepackage{esvect}

\usepackage{stackengine,scalerel,amssymb}

\DeclareMathOperator*{\argmin}{min}

\usepackage{forest}
\forestset{my tree/.style = {
    for tree={
        inner sep=0.075cm, 
        l sep=1mm,
        s sep=1mm,
        grow'=0,
        child anchor=west,parent anchor=south,
        anchor=west,calign=first,
        edge path={
          \noexpand\path [draw, \forestoption{edge}]
          (!u.south west) +(7.5pt,0) |- node{} (.child anchor)\forestoption{edge label};
        },
        before typesetting nodes={
          if n=1
            {insert before={[,phantom]} }
            {}
        },
        fit=band,
        before computing xy={l=15pt},
    }
}}

\usepackage{xcolor,colortbl}
\usepackage{array}
\usepackage{arydshln}
\algnewcommand{\LineComment}[1]{\State{\textcolor{blue}{\(\triangleright\) #1}}}

\setcopyright{acmlicensed}
\copyrightyear{2018}
\acmYear{2018}
\acmDOI{XXXXXXX.XXXXXXX}
\acmConference[Conference acronym 'XX]{Make sure to enter the correct
  conference title from your rights confirmation emai}{June 03--05,
  2018}{Woodstock, NY}
\acmISBN{978-1-4503-XXXX-X/18/06}

\begin{document}

\title{Taxonomy Tree Generation from Citation Graph}

\author{Yuntong Hu}
\email{yuntong.hu@emory.edu}
\affiliation{%
  \institution{Emory University}
  \city{Atlanta}
  \state{GA}
  \country{USA}
}

\author{Zhuofeng Li}
\email{zhuofengli12345@gmail.com}
\affiliation{%
  \institution{Shanghai University}
  \city{Shanghai}
  \country{China}
}

\author{Zheng Zhang}
\email{zheng.zhang@emory.edu}
\affiliation{%
  \institution{Emory University}
  \city{Atlanta}
  \state{GA}
  \country{USA}
}

\author{Chen Ling}
\email{chen.ling@emory.edu}
\affiliation{%
  \institution{Emory University}
  \city{Atlanta}
  \state{GA}
  \country{USA}
}

\author{Raasikh Kanjiani}
\email{raasikh.kanjiani@emory.edu}
\affiliation{%
  \institution{Emory University}
  \city{Atlanta}
  \state{GA}
  \country{USA}
}

\author{Boxin Zhao}
\email{zinc.zhao@emory.edu}
\affiliation{%
  \institution{Emory University}
  \city{Atlanta}
  \state{GA}
  \country{USA}
}

\author{Liang Zhao}
\email{liang.zhao@emory.edu}
\affiliation{%
  \institution{Emory University}
  \city{Atlanta}
  \state{GA}
  \country{USA}
}

\renewcommand{\shortauthors}{Yuntong et al.}

\begin{abstract}
Constructing taxonomies from citation graphs is essential for organizing scientific knowledge, facilitating literature reviews, and identifying emerging research trends. However, manual taxonomy construction is labor-intensive, time-consuming, and prone to human biases, often overlooking pivotal but less-cited papers. In this paper, to enable automatic hierarchical taxonomy generation from citation graphs, we propose HiGTL (Hierarchical Graph Taxonomy Learning), a novel end-to-end framework guided by human-provided instructions or preferred topics. Specifically, we propose a hierarchical citation graph clustering method that recursively groups related papers based on both textual content and citation structure, ensuring semantically meaningful and structurally coherent clusters. Additionally, we develop a novel taxonomy node verbalization strategy that iteratively generates central concepts for each cluster, leveraging a pre-trained large language model (LLM) to maintain semantic consistency across hierarchical levels. To further enhance performance, we design a joint optimization framework that fine-tunes both the clustering and concept generation modules, aligning structural accuracy with the quality of generated taxonomies. Extensive experiments demonstrate that HiGTL effectively produces coherent, high-quality taxonomies.
\end{abstract}



\received{20 February 2007}
\received[revised]{12 March 2009}
\received[accepted]{5 June 2009}

\maketitle

\section{Introduction}
\begin{figure}[t!]
  \centering
  \includegraphics[width=0.45\textwidth]{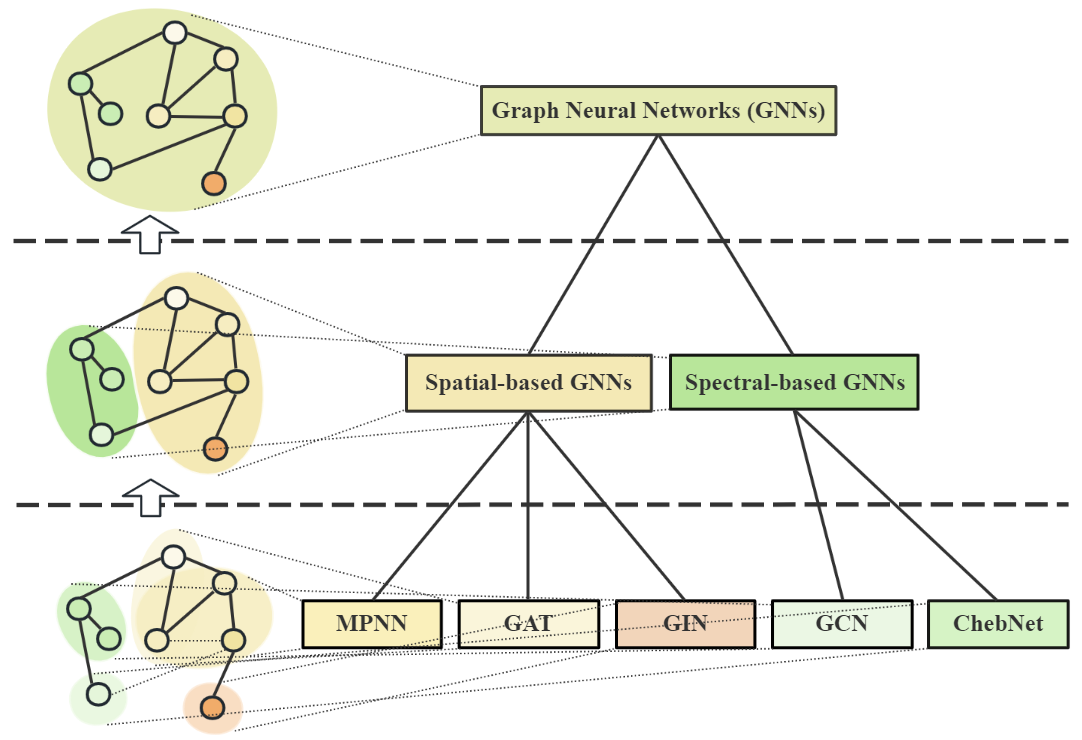}
  \caption{The citation graph and its taxonomy exhibit a clear hierarchical mapping.}
  \label{problem}
\vspace{-16pt}
\end{figure}

Citation graphs represent the citation relationships within a collection of documents (e.g., research papers, patents, etc.), capturing the structure of scholarly communication and illustrating how ideas evolve and influence one another. A taxonomy of citation graphs organizes research topics hierarchically, grouping related papers into clusters that form broader categories (as shown in \hyperref[problem]{Figure} \ref{problem}). This tree-like structure highlights the connections between specific subtopics and overarching fields, offering an interpretable overview of the research landscape. Such taxonomies are typically substantial backbones of subsequent activities such as literature reviews \cite{shen2018web}, meta-analysis and knowledge discovery \cite{hearst1992automatic, liu2012automatic}, and research trend identification \cite{ponzetto2010knowledge}, which are vital to the scientific community. However, manual taxonomy construction is labor-intensive, time-consuming, and prone to human bias and comprehensiveness, which hence struggle to keep pace with the rapid growth of publications. In natural language processing, there are techniques called existing taxonomy learning which focuses on deriving semantic relationships from linguistic patterns in texts~\cite{hearst1992automatic, pantel2004automatically, snow2006semantic, suchanek2007yago}. They are designed for plain text corpora rather than text graphs and hence cannot handle patterns like connectivities and communities of documents, which are crucial for taxonomy generation from citation graphs \cite{liu2012automatic, de2013domain, dietz2012taxolearn, zhang2018taxogen}. In this paper, we focus on a compelling problem:

\begin{tcolorbox}[colback=gray!10, colframe=white, boxrule=0pt, arc=4pt, width=\linewidth, boxsep=3pt, left=2pt, right=2pt, center title]
\centering
\textit{Can we automatically generate taxonomies from citation graphs?}
\end{tcolorbox}

Generating taxonomy trees from citation graphs, though important, encompasses significant and unique major challenges to be addressed:
\textit{(1) How do graph topology and texts jointly decide taxonomy?} 
Citation graphs contain both 1) the citing relationships among papers with similar topics of interest and 2) the rich textual content of each paper. Effectively integrating these two modalities is crucial because relying solely on graph topology may overlook semantic nuances, while focusing only on text may ignore the technique evolution and contextual relevance indicated through citation links.
\textit{(2)  How are the taxonomy nodes verbalized?} 
Verbalizing taxonomy nodes involves generating concise, accurate labels that capture the essence of each cluster while maintaining semantic coherence across hierarchical levels. This is challenging because clusters often span diverse or overlapping topics, requiring the model to balance specificity and generalization to produce informative and contextually appropriate labels.
\textit{(3) How to automatically learn the whole taxonomy generator from data?} 
 Transforming from citation graphs to taxonomy is a nontrivial tasks that require a highly expressive model for both taxonomy tree construction and taxonomy node verbalization. How to train such a model from limited data is challenging in both the optimization and learning aspects.

To pursue our goal and address these challenges, we propose \textbf{Citation Graph to Taxonomy Transformation (CGT)}, a novel framework that automatically extracts taxonomies from citation graphs. First, to enable meaningful hierarchical citation graph clustering, we propose a novel method that recursively decomposes the citation graph into multiple levels by grouping papers based on both textual content and citation link topology. At each level, nodes of lower-level topics are clustered into ``supernodes'' of higher-level topics, while feature aggregation maintains semantic coherence, facilitating the extraction of structured taxonomies. Second, to verbalize the nodes in taxonomy, we introduce a hierarchical concept abstraction strategy that iteratively generates central concepts for each cluster. The method synergizes cluster-level graph embeddings and paper-level information to produce consistent, semantically rich taxonomies aligned with user-preferred topics by prompting. Finally, to ensure clusters are structurally meaningful and semantically coherent, we jointly optimize all the parameters of our CGT through a customized training strategy consisting of pre-training and fine-tuning. To thoroughly evaluate the effectiveness of our method, we collected 518 citation graphs, each corresponding to a citation graph of a high-quality, human-written literature review in the computer science domain. Extensive experiments demonstrate that our model effectively generates high-quality taxonomies.

In summary, our primary contributions are as follows:
\begin{itemize}[left=0pt]
\item We introduce a novel end-to-end framework that automatically generates taxonomies from citation graphs, guided by human-provided central topics to enhance relevance.
\item We propose a hierarchical citation graph clustering approach for citation graphs that integrates textual content and structural information, enabling the formation of semantically meaningful and structurally coherent clusters.
\item We develop a taxonomy node verbalization method that iteratively generates central concepts for clusters at each level, ensuring semantic consistency throughout the hierarchy.
\item We design a novel optimization framework that jointly fine-tunes hierarchical citation graph clustering and taxonomy node verbalizer, improving both structural accuracy and the coherence of the generated taxonomies.
\end{itemize}

\section{Related Work}\label{sec:related work}

\subsection{Taxonomy Learning} 
Taxonomy learning \cite{hearst1992automatic,pantel2006espresso,suchanek2006combining} aims to construct hierarchical structures that capture semantic relationships within domain-specific corpora. Its objective is to identify and organize conceptual relationships while maximizing semantic coherence and relevance. Existing approaches can be categorized into four main paradigms: pattern-based methods \cite{suchanek2006combining,ponzetto2011taxonomy,rios2013learning}, clustering-based techniques \cite{liu2012automatic,dietz2012taxolearn}, statistical frameworks \cite{wang2009probabilistic,diederich2007semantic}, and graph-based approaches \cite{kozareva2010semi,velardi2013ontolearn,kang2015taxofinder}.
While these methods have advanced our understanding of taxonomy construction, they predominantly focus on unstructured database \cite{huang2019unsupervised,qiu2018automatic}, limiting their applicability to more complex structured databases. Recent advances have extended taxonomy learning to graph-structured data, particularly knowledge graphs \cite{martel2021taxonomy}. However, these approaches are primarily designed for knowledge graphs where nodes represent atomic concepts or entities. This makes them insufficient for real-world graph databases like citation networks, where nodes contain rich textual content.

\subsection{Hierarchical Graph Clustering} 
The primary goal of hierarchical graph clustering is to construct a hierarchy of clusters that effectively captures the underlying multi-scale structure of a graph. \citeauthor{flake2004graph} introduced a minimum s-t-cuts-based hierarchical graph clustering method, guaranteeing high-quality partitions. \citeauthor{doll2011fully} extended this approach by allowing arbitrary minimum s-t-cuts, enabling more flexible clustering hierarchies. While these methods provide strong theoretical guarantees, they struggle with scalability on large-scale graphs.
To improve efficiency, density-based methods have been explored \cite{schlitter2014dengraph, szilagyi2014fast}, grouping nodes based on high local density, making them robust to variations in graph topology. Modularity-based methods have also been successful. \citeauthor{zeng2015parallel} proposed a parallel hierarchical graph clustering algorithm leveraging distributed memory architectures and a divide-and-conquer strategy for large-scale graphs. These methods remain widely used in unsupervised community detection \cite{bonald2018hierarchical, zeng2015parallel}.
To improve scalability, minimum spanning tree (MST)-based clustering methods, such as Affinity Clustering \cite{elmahdy2020matrix} and Hierarchical Spectral Clustering \cite{singh2025end}, have been proposed to recursively partition graphs.


\subsection{Parameter-Efficient Fine-Tuning (PEFT)} 
Full fine-tuning of domain-specific large pre-trained models is resource-intensive, while parameter-efficient fine-tuning (PEFT) improves performance in specific domains by updating only a subset of parameters \cite{ding2023parameter,han2024parameter}. Key techniques in PEFT include Adapters, Low-Rank Adaptation (LoRA) \cite{hu2021lora}, and Prompt Tuning. Adapters are small neural networks inserted into the layers of a pre-trained model to modify its behavior for specific tasks \cite{houlsby2019parameter,pfeiffer2020adapterfusion}. Instead of updating the entire model, only the adapter parameters are updated, making the process computationally efficient. Adapters have also been explored in applying PEFT to graph-based language models \cite{chai2023graphllm,perozzi2024let,liu2024can}. LoRA introduces low-rank matrices into transformer layers, enabling fine-tuning of only a small subset of the model’s parameters while preserving overall performance. Prompt Tuning involves learning task-specific prompts that guide the model to perform various tasks without altering its underlying parameters \cite{lester2021power,li2021prefix,liu2023pre}. The pre-trained model remains fixed, with only the input prompts being optimized for the specific task.

\begin{figure*}[t!]
  \centering
  \includegraphics[width=\textwidth]{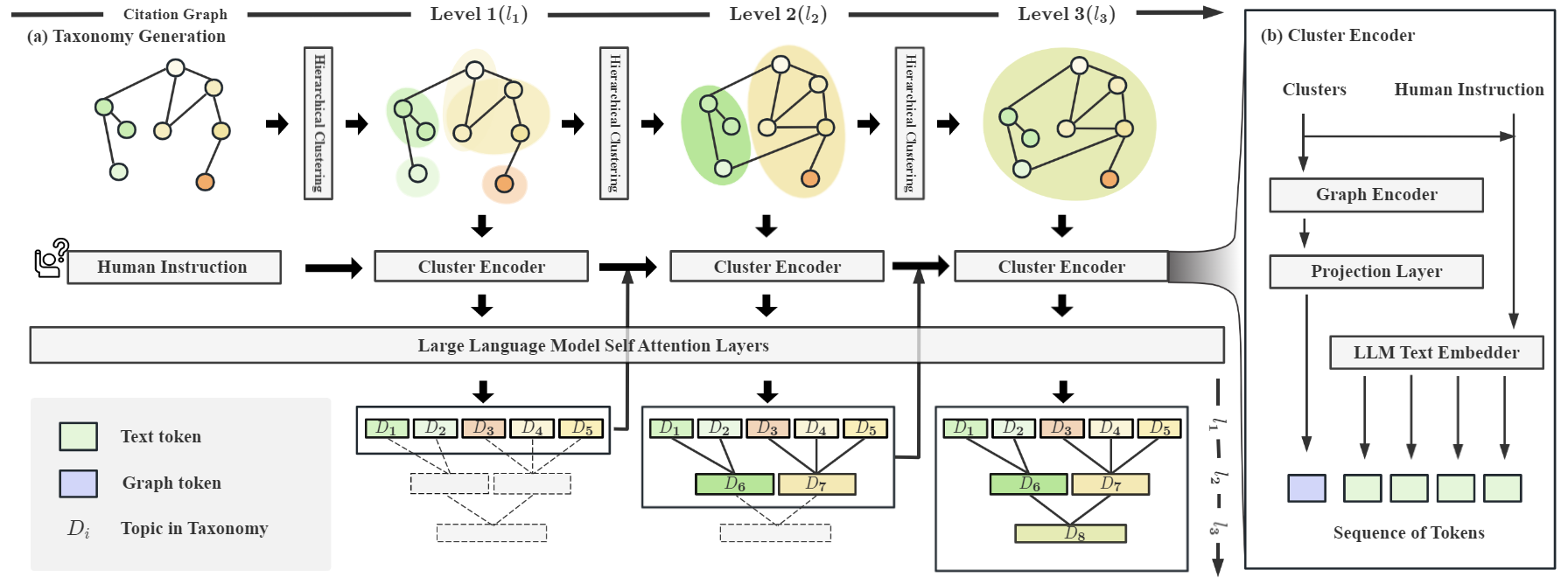}
  \caption{Overall framework of proposed hierarchical taxonomy generation.}
  \label{Flow}
  \vspace{-7pt}
\end{figure*}

\section{Problem Formalization}\label{sec:problem}

\vspace{2pt}
\noindent \textbf{Citation Graph.}\quad A citation graph is a directed graph defined as \( \mathcal{G} = (V, E, \{D_v\}_{v \in V}) \), where \( V \) is the set of nodes (papers), \( E \subseteq V \times V \) represents the directed edges (citation relationships), and \( D_v = [w_1, w_2, \dots, w_n] \) denotes the tokenized textual content of paper \( v \), where \( w_i \in \mathcal{W} \), \( \mathcal{W} \) represents the token vocabulary, and \( n \) is the number of tokens in \( D_v \).

\vspace{2pt}
\noindent \textbf{Taxonomy Tree.}\quad A taxonomy tree, \( \mathcal{T} = (U, R, \{D_u\}_{u \in U}) \), is a rooted tree where \( U \) is the set of nodes representing topics or subtopics, \( R \subseteq U \times U \) denotes directed edges capturing hierarchical relationships, and \( D_u = [w_1, w_2, \dots, w_k] \) represents the tokenized textual content associated with each topic \( u \), where \( w_i \in \mathcal{W} \) and \( k \) is the number of tokens in \( D_u \). The taxonomy tree outlines relationships among different topics in the citation graph, with each topic corresponding to a  paper cluster.

\vspace{2pt}
\noindent \textbf{Taxonomy Tree Generation from Citation Graph.}\quad 
The goal of taxonomy generation is to learn a mapping function \( f \) that constructs an optimal taxonomy tree from the citation graph for a specific topic $q$, i.e., \( f: \mathcal{G},q \rightarrow \mathcal{T^*} \).
To achieve this goal, three challenges must be addressed:
\begin{itemize}[left=0pt]
    \item \textbf{Graph-to-Taxonomy Transformation.} Citation graphs exhibit complex citation relationships, often allowing papers to belong to multiple topics, whereas taxonomy trees impose a strict hierarchical structure with a single-parent constraint.
    \item \textbf{Semantic Coherence in Taxonomy Generation.} Citation graphs contain extensive textual data with redundancy across papers. Constructing a concise, coherent taxonomy requires effective topic partitioning while preserving semantic integrity.
    \item \textbf{The Learning of Taxonomy Generation.} 
    The optimization must ensure structural validity and semantic coherence by jointly refining tree structure and topic generation, aligning fine-grained clusters with higher-level generated concepts.
\end{itemize}


\section{Methodology}\label{sec:method}

\subsection{Overview}
Given a citation graph \( \mathcal{G} = (V, E, \{D_v\}_{v \in V}) \), taxonomy generation for $\mathcal{G}$ involves both structural organization and concept compression. Formally, let a hierarchy induction function \( g: (V, E) \to (U, R) \) define the hierarchical organization of topics, and let a concept abstraction function \( h: \{D_v\}_{v \in V} \to \{D_u\}_{u \in U} \) compress and summarize textual content at each taxonomy level. We argue that these mappings are inherently interdependent, forming an entangled transformation expressed as \( f = h \circ g \) (\hyperref[Flow]{Figure} \ref{Flow}\hyperref[Flow]{(a))}, which decomposes the taxonomy generation process for a citation graph into two interlinked subproblems: \hyperref[sec:clustering]{\textbf{Hierarchical Citation Graph Clustering}} and \hyperref[sec:tree_generation]{\textbf{Taxonomy Node Verbalization}}.
A well-designed hierarchy induction function \( g \) establishes the relationship between the citation graph and its taxonomy. Subsequently, an effective concept abstraction function \( h \) extracts key concepts and topics from the citation graph. To achieve this decomposition and address the challenges at the end of
\hyperref[sec:problem]{Section}~\ref{sec:problem}, we propose \textbf{\underline{Hi}erarchical \underline{G}raph \underline{T}axonomy \underline{L}earning (HiGTL)} for citation graph:
\begin{itemize}[left=0pt]
    \item In \hyperref[sec:clustering]{Section}~\ref{sec:clustering}, we propose a novel hierarchical graph clustering approach as the hierarchy induction function to address the \textbf{Graph-to-Taxonomy Transformation} challenge. We hierarchically categorize papers in a citation network, ensuring that each cluster at a given level corresponds to a topic in the taxonomy tree at the same level.
    \item In \hyperref[sec:tree_generation]{Section}~\ref{sec:tree_generation}, to achieve taxonomy node verbalization and address the challenge of \textbf{Semantic Coherence in Taxonomy Generation}, we propose an iterative graph-taxonomy generation approach—a bottom-up method that identifies the central topic of each cluster at every hierarchical level while ensuring conceptual alignment across levels.
    \item In \hyperref[sec:training]{Section}~\ref{sec:training}, we propose a novel two-phase optimization strategy for \textbf{the Learning of Taxonomy Generation}, enabling end-to-end joint training of graph clustering and taxonomy generation. The process begins with pretraining the hierarchical citation graph clustering module using discrete clustering labels, followed by fine-tuning both the clustering and generation modules with continuous labels representing topic content.
\end{itemize}

\subsection{Hierarchical Citation Graph Clustering} \label{sec:clustering}

Given a citation graph \( \mathcal{G} \), let \( \{\mathcal{G}^{(l)} = (V^{(l)}, E^{(l)}, X^{(l)})\}_{l=1}^L \) represent its hierarchical decomposition into a sequence of (hyper-)graphs across \( L \) levels, where \( V^{(l)} \) are the nodes, \( E^{(l)} \) are the edges, and \( X^{(l)} \) are the node representations at level \( l \). The recursive updates are defined as:
\begin{align}
    &(V^{(l+1)}, E^{(l+1)}) = \text{CLU}(V^{(l)}, E^{(l)}, X^{(l)}), \\
    &X^{(l+1)} = \text{AGG}(\text{CLU}(V^{(l)}, E^{(l)}, X^{(l)})),
\end{align}
where \( \text{CLU}(\cdot) \) clusters nodes in \( \mathcal{G}^{(l)} \) into hyper-nodes for \( \mathcal{G}^{(l+1)} \) and forms edges between them (elaborated in Section~\ref{sec: clustering operator}), while \( \text{AGG}(\cdot) \) computes features for hyper-nodes in \( V^{(l+1)} \) (elaborated in Section~\ref{sec: Aggregation Operator}). The sequence is initialized from the citation graph \( \mathcal{G}^{(1)} = \mathcal{G} \), with node features \( X^{(1)} = \text{LM}(\{D_v\}_{v \in V}) \) generated using pre-trained dense language models.

The hierarchical citation graph clustering and aggregation operations collectively define the hierarchy induction function \( g: (V, E) \to (U, R) \). This hierarchical citation graph clustering partitions the node set \( V^{(l)} \) into clusters \( \mathcal{C}^{(l)} \), with each hyper-node \( u \in V^{(l+1)} \) corresponding to a unique topic \( u \in U \). The edge set \( R \) is directly induced from the evolving hierarchical aggregation across hierarchical levels.

\vspace{2pt}
\subsubsection{Clustering Operator \( \text{CLU}(\cdot) \).} \label{sec: clustering operator}\quad
The clustering function partitions nodes in \( V^{(l)} \) into clusters based on structural and feature similarities, treating each cluster as a hyper-node in \( V^{(l+1)} \). 
Edges \( E^{(l+1)} \) are formed between hyper-nodes if any nodes in their corresponding clusters were connected in \( E^{(l)} \):
\[
E^{(l+1)} = \{(c^{(l)}_i, c^{(l)}_j) \mid \exists (u, v) \in E^{(l)}, \ u \in c^{(l)}_i, \ v \in c^{(l)}_j \}.
\]
where \( c^{(l)}_i \) and \( c^{(l)}_j \) represent node clusters at level \( l \). We next introduce our strategy for extracting clusters at each hierarchical level.

Given \( \mathcal{G}^{(l)} = (V^{(l)}, E^{(l)}, X^{(l)}) \), we apply a graph encoder, such as GNN$_{\theta}$, which aggregates information from neighboring nodes, updating node embeddings. We then compute the probability \( \hat{p}^{(l)}_{uv} \) of nodes \( u, v \in V^{(l)} \) belonging to the same cluster using MLP$_{\phi^{(l)}}$ with a softmax transformation, taking the concatenated embeddings \( [h_u; h_v] \) as input.
We then calculate the node density \( \hat{d}_u \) to measure the similarity-weighted proportion of same-cluster nodes in its neighborhood:
\begin{equation}
    \hat{d}_u = \frac{1}{|\mathcal{N}(u)|}\sum_{k \in \mathcal{N}(u)} \hat{p}^{(l)}_{uk}\cdot \frac{h_u \cdot h_k}{\|h_u\|\|h_k\|}.
\end{equation}
where \( \mathcal{N}(u) \) indicates the neighboring nodes of node \( u \). The density assesses how densely connected each node is within its local neighborhood. High-density nodes are more likely to be in the core of a cluster, whereas low-density nodes are more likely to be in ambiguous regions between clusters. \( \hat{d}_u = 0 \) if \( u \) is isolated.

To balance fine-grained topic relationships with a well-structured taxonomy, we employ soft clustering at the first level, allowing topic overlap to reflect the interdisciplinary nature of research, and progressively transition to hard clustering at higher levels to enforce distinct, non-overlapping categories in the taxonomy hierarchy.

\textit{At the base level} (\( l=1 \)), given $\hat{p}^{(l)}_{uv}$, $\hat{d}_{u}$, $\hat{d}_{v}$, and a pre-defined edge connection threshold $p_{\tau}$, we generate the candidate cluster as:
\begin{equation}
    c^{(l)}_i = \{u\} \cup \{v \mid \hat{d}_u < \hat{d}_v \ \text{and} \ \hat{p}^{(l)}_{uv} > p_{\tau}\}.
\end{equation}
Nodes with higher density may belong to multiple clusters, reflecting the interdisciplinary nature of research. Final clusters are formed by merging candidate clusters:
\begin{equation}
    \mathcal{C}^{(l)} = \{c^{(l)}_i \mid c^{(l)}_i \not\subset c^{(l)}_j, \ |c^{(l)}_i| \neq 1\}.
\end{equation}

\textit{At higher levels} (\( l > 1 \)), clustering transitions to hard clustering to enforce distinct, non-overlapping categories. Edges between nodes are formed based on the strongest connection probabilities:
\begin{equation}
    \mathcal{E} = \{(u, v) \mid \arg\max_{v \neq u \in V} \hat{p}^{(l)}_{uv}\}, \quad u \in V^{(l)}.
\end{equation}
After traversing all nodes, connected components defined by \( \mathcal{E} \) form disjoint clusters $\mathcal{C}^{(l)} $, ensuring clear topic boundaries within the taxonomy hierarchy.

\vspace{2pt}
\subsubsection{Aggregation Operator $\text{AGG}(\cdot)$.}\label{sec: Aggregation Operator}\quad
The aggregation function computes the feature representation for each hyper-node in \( V^{(l+1)} \). Specifically, the feature of a hyper-node is derived from both the average and the most representative node features within its corresponding cluster:
\begin{equation}
    x_u = \frac{1}{|c^{(l)}_i|} \sum_{z \in c^{(l)}_i} h_z + h_k, \quad \text{where} \quad k = \arg\max_{z \in c^{(l)}_i} \hat{d}_z.
\end{equation}
Here, \( h_k \) corresponds to the node with the highest density in cluster \( c^{(l)}_i \), ensuring that the hyper-node captures both general and distinctive characteristics of the cluster. This aggregated feature set \( X^{(l+1)} \) serves as the input for the next hierarchical level.

\vspace{2pt}
\subsubsection{Hierarchical Citation Graph Clustering Objective.}\quad
Since paper-level annotations are the most granular and reliable supervision, leveraging them indirectly optimizes higher-level clusters, preserving semantic integrity and aligning the taxonomy with real-world research structures without relying on hypothetical hypernode labels.
Moreover, papers consistently clustered together across multiple hierarchical levels should be closer in the embedding space than those grouped only at lower levels or not at all, reflecting stronger and more persistent semantic relationships.
Therefore, we design the objective function to operate on paper-level labels as:
\begin{equation}
    \mathcal{L}_{\text{HiCluster}} = \mathcal{L}_{\text{cluster}} + \alpha\cdot \mathcal{L}_{\text{HiMulCon}},
\end{equation}
where the first term, \( \mathcal{L}_{\text{cluster}} \), optimizes clustering by grouping papers with structural and semantic similarities at each hierarchical level. The second term, \( \mathcal{L}_{\text{HiMulCon}} \), is a hierarchical contrastive loss that enforces multi-level consistency by bringing together nodes that persist in the same cluster across levels and pushing apart those that diverge, ensuring the embeddings reflect the taxonomy's hierarchical structure. $\alpha$ controls the contribution from the contrastive loss. Specifically, the clustering loss term is defined as follows,
\begin{equation}
    \mathcal{L}_{\text{cluster}} = \sum_{l=1}^L \frac{-1}{|E|}\sum_{(u,v)\in E} q_{uv}^{(l)} \log \hat{p}_{uv}^{(l)} + (1-q_{uv}^{(l)}) \log (1-\hat{p}_{uv}^{(l)})
\end{equation}
where \( q_{uv}^{(l)} = 1 \) if \( u,v\in V\) belong to the same cluster at level \( l \), and \( q_{uv}^{(l)} = 0 \) otherwise. This is computed over the edge set \( E \) rather than all node pairs, as we empirically found it reduces computational costs without significantly affecting performance. 

To ensure that papers consistently assigned to the same cluster across multiple levels are positioned closer in the embedding space, the second term is formulated as a hierarchical contrastive loss as:

\begin{footnotesize}
\begin{equation}
    \mathcal{L}_{\text{HiMulCon}} = \frac{1}{|L|} \sum_{l=1}^L \sum_{u \in V} \frac{-\delta_l}{|S^{(l)}_u|} \sum_{k \in S^{(l)}_u}\log \frac{\exp(\text{sim}(h_u, h_{k}) / \tau)}{\sum_{v \in V \setminus u} \exp(\text{sim}(h_u, h_v) / \tau)}
\end{equation}
\end{footnotesize}

\noindent where \( S^{(l)}_u \) represents the set of positive samples for node \( u \) at level \( l \), i.e., nodes that belong to the same cluster. \( \delta_l \) is a weighting factor that reflects the importance of each level for the loss, and \( \tau \) is the temperature parameter. $L$ represents the total number of hierarchical levels.

\subsection{Hierarchical Taxonomy Node Verbalization} \label{sec:tree_generation}
A citation graph \( \mathcal{G} = (V, E, \{D_v\}_{v \in V}) \) reveals local research communities that expand into larger clusters through hierarchical analysis. Its structure supports a bottom-up taxonomy, where denser subgraphs form the basis for broader categories. 
After hierarchical citation graph clustering, each cluster \( c^{(l)} \in \mathcal{C}^{(l)} \) corresponds to a topic \( u \in U \) in the taxonomy \( \mathcal{T} \). We now introduce an iterative generation method to achieve the mapping \( h: \{D_v\}_{v \in V} \to \{D_u\}_{u \in U} \) for summarizing the central concept of each cluster.
Specifically, we propose an iterative graph-taxonomy generation strategy that refines fine-grained topics and abstracts them into higher-level concepts, ensuring semantic consistency across hierarchy levels. 

\subsubsection{Iterative Generation}  


Given the hierarchical decomposition \( \mathcal{G}^{(l)} = (V^{(l)}, E^{(l)}, X^{(l)}) \) of a citation graph \( \mathcal{G} \) at level \( l \) and its corresponding clusters \( \mathcal{C}^{(l)} \), we generate the central concept for each cluster.
To generate a meaningful taxonomy for a specific topic \( q \), we employ a pre-trained LLM$_\Theta$ as the backbone for concept abstraction, leveraging its extensive text processing capabilities. For each cluster \( c^{(l)} \in \mathcal{C}^{(l)} \), the probability distribution of its generated central concept is defined as:
\begin{equation}\label{gen}
    p_{\Theta, \Phi} (Y_u^{(l+1)} \mid c^{(l)}, q) = \prod_{i=1}^{r} p_{\Theta, \Phi} (y_i \mid y_{<i}, c^{(l)}, q),
\end{equation}
where \( Y_u^{(l+1)} \) denotes the token sequence of central concept of cluster \( c^{(l)} \in \mathcal{C}^{(l)} \), corresponding to the hyper-node \( u \in V^{(l+1)} \). \( q \) represents the instruction prompt reflecting user preferences, and \( y_{<i} \) denotes the prefix tokens. $\Phi$ indicates the parameters of a projector that maps the graph embedding into the LLM text embedding space.

Each cluster at level \( l \) represents a subgraph of \( \mathcal{G}^{(l)} \). To capture interactions between hyper-nodes for central concept generation, we apply a projector MLP$_\Phi$ to map graph embeddings into the LLM text space as graph tokens, where graph embeddings are produced using the same GNN$_{\theta}$ in the hierarchical citation graph clustering process. While hierarchical aggregation retains cluster features in hyper-nodes, information loss from individual papers is inevitable as the hierarchy deepens. To address this, we also incorporate paper-level information from the base graph, as each hyper-node at higher levels can be coarsened to its corresponding set of base-level papers. Formally, the cluster \( c^{(l)} \) is encoded as:
\begin{equation}
    \mathbf{h}_{c} = \text{MLP}_\Theta(\text{GNN}_\theta(c^{(l)})) \in \mathbb{R}^{d_{\text{LLM}}}
\end{equation}
where \( d_{\text{LLM}} \) denotes the LLM text embedding dimension. The associated papers and instructions are encoded using the LLM text embedder as:
\begin{align}
    &\mathbf{h}_{\pi} = \text{TextEmbedder}(\{D_v\}_{v \in \pi_c^{(l)}}) \in \mathbb{R}^{N_\pi\times d_{\text{LLM}}},\\
    &\mathbf{h}_{q} = \text{TextEmbedder}(q) \in \mathbb{R}^{N_q\times d_{\text{LLM}}},
\end{align}
where \(N_\pi\), \(N_q\) indicate the length of token sequences, \( \pi_c^{(l)} \) represents the set of base-level papers corresponding to the coarsened \( c^{(l)} \). Finally, the probability distribution in \hyperref[gen]{Equation} (\ref{gen}) is as,
\begin{equation}
    p_{\Theta, \Phi} (Y_u^{(l+1)} \mid c^{(l)}, q) = \prod_{i=1}^{r} p_{\Theta, \Phi} (y_i \mid y_{<i}, [\mathbf{h}_c; \mathbf{h}_\pi; \mathbf{h}_q]),
\end{equation}
where  $[~;~]$ indicates the concatenation of token embeddings before feeding them through transformer layers of LLM$_\Theta$.

\subsubsection{Hierarchical Generation Objective}  The goal of taxonomy generation is to maximize the likelihood of generating coherent central concepts for all clusters across the hierarchical citation graph, thereby constructing the entire taxonomy tree. Formally, the objective is defined as:
\begin{equation}
    \mathcal{L}_{\text{Gen}} = \sum_{l=1}^{L} \sum_{c^{(l)} \in \mathcal{C}^{(l)}} \log p_{\Theta, \Phi}(Y_u^{(l+1)} \mid c^{(l)}, q).
\end{equation}
By maximizing this hierarchical likelihood, the model generates semantically consistent and contextually relevant topics, ensuring coherence throughout the taxonomy tree.

\subsection{Optimization for CGT}\label{sec:training}
We jointly train the hierarchical citation graph clustering model and the taxonomy node verbalizer to ensure that the hierarchical clusters formed are meaningful from a textual perspective, and that the topics generated are coherent with the structural information embedded in the citation network. The final objective function is given as,
\begin{equation}
    \argmin_{\Theta, \Phi, \theta, \phi} \mathcal{L} =  \mathcal{L}_{\text{Gen}}(\Theta,\Phi) + \lambda \cdot\mathcal{L}_{\text{HiCluster}}(\theta,\phi),
\end{equation}
where $\lambda$ is a pre-defined parameter, $\mathcal{L}_{\text{HiCluster}}$ is the loss function for hierarchical citation graph clustering, and $\mathcal{L}_{\text{Gen}}$ is the loss function for taxonomy node verbalization. Since only a small portion of the citation graph has been labeled for hierarchical citation graphclustering (as it is challenging to collect labels that indicate which hierarchical citation graph cluster the conferences in a literature review belong to), and the learning dynamics of GNNs and LLMs differ significantly, directly training both models simultaneously can result in a complex and unstable optimization process. This leads to the GNN struggling to learn meaningful clusters early on, which subsequently hinders the LLM’s ability to generate coherent topics. Therefore, we pre-train the hierarchical citation graph clustering module to simplify the optimization process for the LLM, allowing the LLM to focus solely on content generation. 

After pre-training the hierarchical clustering module \( (\theta, \phi) \), we then jointly fine-tune them and the concept generator \((\Theta, \Phi) \) to generate the central topic for each cluster.
The fine-tuning is conducted using Low-Rank Adaptation (LoRA) \cite{hu2021lora}, with \( \text{MLP}_{\Phi} \) serving as an adapter for graphs, enabling efficient adjustment of both models.

\section{Experiment}\label{sec:exp}
To demonstrate the effectiveness of the proposed taxonomy learning method, we evaluate the quality of the generated taxonomies and further assess the model's ability to guide literature review generation.
We first introduce the experimental settings in \hyperref[sec:setup]{Section} \ref{sec:setup}. Next, we present the performance of HiGTL on citation graphs in the computer science domain in \hyperref[sec:mainresults]{Section}~\ref{sec:mainresults}. We then conduct ablation studies to demonstrate the contributions of Hierarchical Citation Graph Clustering and Taxonomy Node Verbalization in \hyperref[sec:ablation]{Section}~\ref{sec:ablation}. Finally, we include a case study to highlight the strengths and limitations of our framework in \hyperref[sec:case]{Section}~\ref{sec:case}.

\subsection{Experiment Setting}\label{sec:setup}
\subsubsection{Datasets} 
We manually collected 518 high-quality literature review articles with clear taxonomies or well-defined structures from arXiv\footnote{\url{https://arxiv.org/}} across various computer science domains, with most published in the past three years. To evaluate taxonomy learning performance, we extracted taxonomy trees from these reviews and gathered the 1-hop citation graph for each review, representing citation relationships among its direct references. Additionally, we collected the 3-hop citation graph for each review to assess literature review generation based on HiGTL-generated taxonomies.
Further details are provided in \hyperref[app:dataset]{Appendix}~\ref{app:dataset}.

\subsubsection{Comparison methods} We compare our model with the following state-of-the-art taxonomy learning methods: 
\textbf{HiExpan} \cite{shen2018hiexpan}, which constructs taxonomies by recursively expanding seed sets using pattern-based and distributional embedding methods, followed by a global optimization module to refine the taxonomy structure for better coherence. 
\textbf{TaxoGen} \cite{zhang2018taxogen}, which generates topic taxonomies by embedding concept terms into a latent space and applying adaptive spherical clustering and local term embeddings to ensure fine-grained semantic distinctions at each hierarchical level. 
\textbf{NetTaxo} \cite{shang2020nettaxo}, which integrates text and network data for taxonomy construction by learning term embeddings from both contexts and employing instance-level motif selection to enhance hierarchical citation graph clustering and topic differentiation. 
Since our model leverages LLMs, we also include \textbf{GPT-4o} and \textbf{Claude-3.5} as baselines for taxonomy generation. Implementation details are provided in \hyperref[app:imp]{Appendix}~\ref{app:imp}.

\subsubsection{Evaluation Metrics} 
To comprehensively evaluate the model's generation quality, we assess outputs from three perspectives: LLMScore, Human Evaluation, and BertScore \cite{zhangbertscore}. \citeauthor{wang2024autosurvey} demonstrated that LLM-based evaluations align closely with human preferences, so we leverage LLMs to rate the generated content. We also employ BertScore to measure the semantic similarity between generated texts and human-written references. However, human evaluation remains essential for accurately capturing human preferences and for cases without available human-written labels, ensuring a thorough assessment of the generated content, especially given LLM hallucinations. Details of the prompts used for LLM evaluation and the instructions for the human evaluation process are provided in \hyperref[app:eval]{Appendix}~\ref{app:eval}.

\subsection{Main Results}\label{sec:mainresults}

\subsubsection{Taxonomy Learning Results}
Results in \hyperref[tab:tax]{Table} \ref{tab:tax} demonstrate that HiGTL outperforms all other models across multiple evaluation metrics, achieving the highest scores in Coverage (0.9357), Structure (0.9413), Relevance (0.8748), and the overall LLM Average (0.9173). It also leads in Human Evaluation metrics with the highest Adequacy (0.7150) and Validity (2.6700), reflecting its superior quality in both automated and human assessments. Additionally, HiGTL achieves the highest BertScore (0.8694), indicating strong alignment with reference taxonomies. Among large language models, GPT-4o shows competitive performance, particularly in Relevance (0.8828), where it slightly edges out HiGTL. However, it falls behind in other areas like Structure and Coverage, as well as human evaluations. Traditional taxonomy learning methods like NetTaxo and TaxoGen consistently lag behind HiGTL and LLM-based approaches due to their lack of ability to capture continuous, nuanced concepts.

\begin{table*}[tb]
\centering
\caption{Results of taxonomy generation. The best performance is in \textbf{Bold}. $\uparrow$ indicates that a higher metric value corresponds to better model performance.}\label{tab:tax}
\begin{tabular}{l|cccc|cc|c}
\hline
\textbf{\multirow{2}*{Model}} &  \multicolumn{4}{c|}{{\fontfamily{pcr}\selectfont LLMScores$\uparrow$}} & \multicolumn{2}{c|}{{\fontfamily{pcr}\selectfont Human Evaluation$\uparrow$}} & \multirow{2}*{\fontfamily{pcr}\selectfont BertScore $\uparrow$} \\ 
 \cline{2-5} \cline{6-7} 
 & Coverage $\uparrow$ & Structure $\uparrow$ & Relevance $\uparrow$ & Average $\uparrow$ & Adequacy$\uparrow$ & Validity$\uparrow$ & \\\hline
\multicolumn{8}{c}{{\fontfamily{pcr}\selectfont Large Language Models}} \\ \hline
\textbf{GPT-4o} & $0.9132_{\pm0.015}$ & $0.8914_{\pm0.018}$ & \textbf{0.8828}$_{\pm0.012}$ & $0.8958$ & 0.6400 & 2.4400 & $0.8376_{\pm0.014}$ \\ \hdashline
\textbf{Claude-3.5} & $0.8821_{\pm0.019}$ & $0.8734_{\pm0.016}$ & $0.8725_{\pm0.014}$ & $0.8760$ & 2.2600 & 2.6000 & $0.8319_{\pm0.013}$ \\\hline
\multicolumn{8}{c}{{\fontfamily{pcr}\selectfont Taxonomy Learning Methods}} \\ \hline
\textbf{HiExpan} & $0.8427_{\pm0.021}$ & $0.8538_{\pm0.018}$ & $0.8219_{\pm0.020}$ & $0.8395$ & 0.4650 & 1.5000 & $0.8145_{\pm0.015}$ \\ \hdashline
\textbf{TaxoGen} & $0.8645_{\pm0.017}$ & $0.8812_{\pm0.014}$ & $0.8456_{\pm0.019}$ & $0.8638$ & 0.5300 & 1.5600 & $0.8249_{\pm0.012}$ \\\hdashline
\textbf{NetTaxo} & $0.8734_{\pm0.016}$ & $0.8721_{\pm0.013}$ & $0.8508_{\pm0.015}$ & $0.8654$ & 0.5150 & 1.7100 & $0.8213_{\pm0.011}$ \\  \hdashline
\rowcolor{gray!20} \textbf{HiGTL} & \textbf{0.9357}$_{\pm0.012}$ & \textbf{0.9413}$_{\pm0.010}$ & \textbf{$0.8748_{\pm0.013}$} & \textbf{0.9173} & \textbf{0.7150} & \textbf{2.6700} & \textbf{0.8694}$_{\pm0.010}$ \\ 
\hline
\end{tabular}
\end{table*}

The potential of taxonomy-guided literature review generation led us to investigation an automatic literature review generation method based on our proposed HiGTL, called \underline{\textbf{Hi}}erarchical Taxonomy-Driven Automatic Literature \underline{\textbf{Review}} Generation (\textbf{HiReview}). The next section presents the results of this literature review generation.

\subsubsection{Review Generation Results}

We compare the reviews generated by our model with those written by human experts, zero-shot LLMs and naive RAG-based LLMs, i.e., GPT-4o and Claude-3.5. Zero-shot LLMs rely solely on their pretrained knowledge to generate literature review content, naive RAG-based LLMs utilize a simple BM25 retriever. The LLM backbone of AutoSurvey and HiReview are both GPT-4o for generating the literature review content. Additionally, we benchmark our model against the state-of-the-art review generator, AutoSurvey \cite{wang2024autosurvey}.
Following AutoSurvey’s scoring criteria, we evaluate the generated reviews based on coverage, structure, and relevance when selecting the best output and calculating the LLMScore. When evaluate the review content, instead of directly scoring the generation, we have LLMs compare the generated content with human-written reviews.

\begin{table*}[tb]
\centering
\caption{Results of literature review generation by pure LLMs, naive RAG-based LLMs, AutoSurvey, and HiReview. The best performance is in \textbf{Bold}. LLM$^\circ$ indicates the LLM is provided with the top-500 relevant papers retrieved by BM25. $\uparrow$ indicates that a higher metric value corresponds to better model performance.}\label{tab:main}
\begin{tabular}{l|cccc|c}
\hline
\textbf{\multirow{2}*{Model}} &  \multicolumn{4}{c|}{{\fontfamily{pcr}\selectfont LLMScores$\uparrow$}} & \multirow{2}*{\fontfamily{pcr}\selectfont BertScore $\uparrow$} \\ 
 \cline{2-5} 
 & Coverage $\uparrow$ & Structure $\uparrow$ & Relevance $\uparrow$ & Average $\uparrow$ & \\\hline
\textbf{Human-written} & $1._{0000}$ & $1._{0000}$ & $1._{0000}$ & $1._{0000}$ & $1._{0000}$ \\\hline
\multicolumn{6}{c}{{\fontfamily{pcr}\selectfont Pure LLMs}} \\ \hline
\textbf{GPT-4o} & 0.7430$_{\pm0.12}$ & 0.8346$_{\pm0.11}$ & 0.8225$_{\pm0.07}$ & 0.8000 & 0.8127$_{\pm0.03}$ \\ \hdashline
\textbf{Claude-3.5} &  0.7224$_{\pm0.09}$ & 0.8116$_{\pm0.14}$ & 0.7948$_{\pm0.09}$ & 0.7763 & 0.8130$_{\pm0.04}$ \\\hline
\multicolumn{6}{c}{{\fontfamily{pcr}\selectfont Naive RAG-based LLMs}} \\ \hline
\textbf{GPT-4o$^\circ$} & 0.8219$_{\pm0.13}$ & 0.8293$_{\pm0.12}$ & 0.8972$_{\pm0.06}$ & 0.8495 & 0.8094$_{\pm0.03}$ \\ \hdashline
\textbf{Claude-3.5$^\circ$} & 0.8339$_{\pm0.11}$ & 0.8215$_{\pm0.13}$ & 0.9051$_{\pm0.05}$ & 0.8535 & 0.8141$_{\pm0.02}$  \\\hline
\multicolumn{6}{c}{{\fontfamily{pcr}\selectfont Auto Review System}} \\ \hline
\textbf{AutoSurvey} & 0.8646$_{\pm0.07}$ & 0.9122$_{\pm0.05}$ & 0.9093$_{\pm0.04}$ & 0.8957 & 0.8256$_{\pm0.02}$ \\  \hdashline
\rowcolor{gray!20} \textbf{HiReview} & \textbf{0.9163}$_{\pm0.03}$ & \textbf{0.9484}$_{\pm0.02}$ & \textbf{0.9428}$_{\pm0.01}$ & \textbf{0.9358} & \textbf{0.8449$_{\pm0.02}$} \\ 
\hline
\end{tabular}
\end{table*}

As shown in \hyperref[tab:main]{Table} \ref{tab:main}, Our method HiReview consistently outperforms the other review generation methods in all metrics. It excels across all LLMScore categories, with notably high structure (0.9484) and relevance (0.9428) scores. AutoSurvey employs a structured methodology that combines retrieval, outline generation, and section drafting, leading to superior content generation compared to naive systems (with average LLMScore of 0.8957).

Pure LLMs and naive RAG-based LLMs struggle with both stability and performance, which makes them unreliable for consistent literature review generation. AutoSurvey reduces this instability through prompt design and multi-output generation, achieving \textit{Structure} ±0.05 and \textit{Relevance} ±0.04—lower deviations than those of pure and naive RAG-based LLMs. HiReview, however, outperforms all other models across all metrics, with consistently low standard deviations. This demonstrates HiReview’s superior stability and consistency in generating high-quality reviews. Its success can be attributed not only to HiReview's use of a graph-context-aware retrieval method but also to the taxonomy tree, which provides hierarchical context for domain-specific concerns within the large language model. An example of a generated literature review section is provided in \hyperref[app:example]{Appendix}~\ref{app:example}.

\subsection{Ablation Study}\label{sec:ablation}

Although we demonstrate the performance of HiReview (taxonomy-then-generation) in terms of the quality of the literature review generated, we will assess the impact of various components on the performance of HiReview. 

\begin{table}[h]
    \centering
    \caption{Ablation study results for HiReview.}
    \label{tab:abl}
    \begin{tabular}{l|ccc}
        \hline
        \textbf{\multirow{2}*{Model}} & \multicolumn{3}{c}{\fontfamily{pcr}\selectfont LLMScores$\uparrow$} \\
        \cline{2-4}
         & Coverage \(\uparrow\) & Structure \(\uparrow\) & Relevance \(\uparrow\) \\ 
        \hline
        \textbf{w/o retrieval}    & 0.6705 & 0.7216 & 0.7073 \\ \hdashline
        \textbf{w/o clustering\(^*\)} & 0.8863 & 0.9261 & 0.9314 \\ \hdashline
        \textbf{w/o taxonomy}     & 0.8612 & 0.8790 & 0.9078 \\
        \hline
    \end{tabular}
\end{table}

\textbf{Impacts of Components.}  As shown in \hyperref[tab:abl]{Table} \ref{tab:abl}, we test three variants of HiReiview mode. \textit{HiReview w/o retrieval} refers to the variant where the graph retrieval module is removed, and all papers in the citation network are used. A significant drop is observed across all metrics, particularly in \textit{Coverage} (0.9163 $\rightarrow$ 0.6705) and \textit{Relevance} (0.9428 $\rightarrow$ 0.7073). This indicates that the inclusion of unrelated papers introduces substantial noise, negatively impacting both the taxonomy tree generation (due to an excess of negative samples in hierarchical citation graph clustering) and content generation (where the noise hinders the creation of precise summaries). As a result, the quality opf generated summaries is even worse than those produced by zero-shot LLMs.

\textit{HiReview w/o clustering$^*$} bypasses the clustering process and directly uses the retrieved papers to generate the taxonomy. Instead of iteratively generating topics at each level, this variant creates the taxonomy in a single step. It is marked with $*$ because the taxonomy node verbalizer in this case is an LLM i.e., GPT-4o, rather than a fine-tuned LLaMA, as the number of taxonomy trees is insufficient for effective fine-tuning. Although this variant performs worse than HiReview, it still delivers competitive performance, outperforming  naive RAG-based LLMs and AutoSurvey. This suggests that the combination of the graph retrieval module and the taxonomy-then-generation paradigm is more effective than naive retrieval-then-generation and outline-then-generation approaches.

\textit{HiReview w/o taxonomy} removes the hierarchical taxonomy tree generation module, instead using paper clusters to prompt the LLM for taxonomy node verbalization and review generation. The absence of a hierarchical taxonomy reduces the model's ability to leverage topic relations across different levels, leading to less organized and relevant content (\textit{Structure}: 0.9484 $\rightarrow$ 0.8790 and \textit{Coverage}: 0.9163 $\rightarrow$ 0.8612). Similar to \textit{HiReview w/o clustering$^*$}, \textit{HiReview w/o taxonomy} does not use fine-tuned LLaMA, and its performance is more degraded. This indicates that when using pure LLM generation methods, generating a hierarchical taxonomy tree to guide content generation significantly enhances the quality of the output. Finally, we can answer remaining questions raised at the beginning of the Experiment Section.

\begin{table}[h!]
    \centering
    \caption{Ablation study results for HiReview.}
    \label{tab:abl}
        \begin{tabular}{l|ccc}
        \hline
        \textbf{Model} &  Level 1 & Level 2 & Average \\\hline
        \textbf{HiCluserting} & 0.7127 & 0.6395 &  0.6761 \\ \hdashline
        \textbf{K-means} & 0.3723 & 0.4201 & 0.3962 \\ \hdashline
        \textbf{LLM clustering} & 0.4296 & 0.4518 & 0.4407  \\ 
        \hline
        \end{tabular}
\end{table}

\begin{figure}[t!]
  \centering
  \begin{minipage}{0.4\textwidth}
    \centering
    \includegraphics[width=\linewidth]{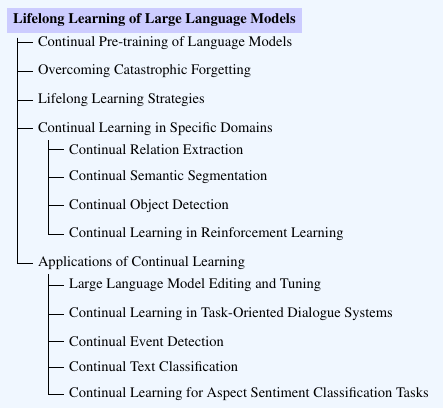}
    \caption{Taxonomy Tree Generated by HiGTL.}
    \label{fig:tree_higtl}
  \end{minipage}%
  \hspace{0.05\textwidth} 
  \begin{minipage}{0.4\textwidth}
    \centering
    \includegraphics[width=\linewidth]{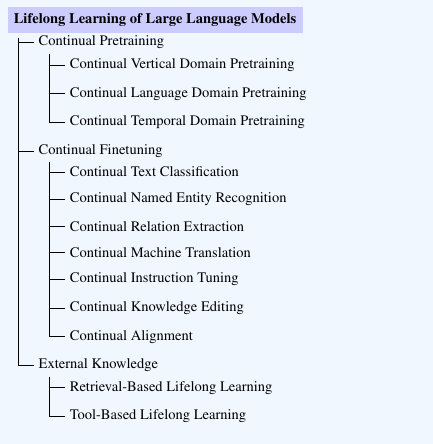}
    \caption{Taxonomy Tree Built by \citeauthor{zheng2024towards}.}
    \label{fig:tree_human}
  \end{minipage}
\end{figure}

We consider two baseline methods for the hierarchical citation graph clustering module: one that utilizes an LLM (i.e., GPT-4o) to cluster papers and another that applies $K$-means, adjusting the number of clusters to represent different levels. However, neither method can be jointly trained with the taxonomy node verbalizer. Even disregarding the training requirement, the hierarchical nature of the literature review's taxonomy tree requires soft clustering at the initial layer and hard clustering at subsequent layers—an issue that no existing work addresses. As shown in \hyperref[tab:cluster]{Table} \ref{tab:cluster}, when considering the clustering task alone, both baselines underperform compared to our hierarchical approach.

As shown in \hyperref[tab:main]{Table} \ref{tab:main}, HiReview, which incorporates a taxonomy tree, outperforms all other models, particularly in \textit{Structure}, achieving a score of 0.9484. In contrast, AutoSurvey, which follows an outline-then-generation approach without hierarchical taxonomy, shows lower scores, such as a \textit{Structure} score of 0.9122. The ablation study further supports this. When the taxonomy is removed, the structure score drops significantly (as in  \hyperref[tab:abl]{Table} \ref{tab:abl}). This demonstrates that the taxonomy tree plays a critical role in organizing and guiding the content generation process, especially when maintaining a clear structure is crucial for a literature review. The taxonomy ensures more coherent and relevant summaries. Without providing the taxonomy tree, the generation loses its hierarchical guidance, leading to less structured and less comprehensive content.

\subsection{Case Study}\label{sec:case}
As shown in \hyperref[fig:tree_higtl]{Figure} \ref{fig:tree_higtl}, the taxonomy generated by HiGTL identifies Continual Text Classification as a key subtopic under Applications of Continual Learning, highlighting its relevance in areas like sentiment analysis, spam detection, and topic categorization. It emphasizes the need for models that can adapt incrementally as new data emerges.
While HiGTL captures broad applications effectively, it may lack the finer distinctions found in human-crafted taxonomies (\hyperref[fig:tree_human]{Figure} \ref{fig:tree_human}), such as domain-specific classifications in legal, medical, or social media contexts. However, this generalized framework ensures semantic coherence and practical usability, providing a solid foundation that human experts can easily refine for tasks like literature review generation.

\section{Conclusion}\label{sec:conclution}
In this paper, we introduced HiGTL, a novel end-to-end framework for automatic taxonomy generation from citation graphs, guided by human-provided instructions or preferred topics. By integrating hierarchical citation graph clustering with taxonomy node verbalization, HiGTL effectively combines the structural relationships inherent in citation networks with the rich textual content of academic papers. Our joint optimization strategy ensures both structural coherence and semantic consistency across hierarchical levels.
Extensive experiments on 518 citation graphs from high-quality literature reviews in computer science demonstrated that HiGTL generates meaningful taxonomies. Furthermore, we extended our framework to literature review generation with HiReview, showcasing its practical utility in guiding the creation of coherent, well-structured reviews. Ablation studies confirmed the critical role of hierarchical clustering and taxonomy-guided generation in enhancing performance, while case studies illustrated the flexibility and usability of the generated taxonomies.
HiGTL offers a scalable, robust solution for organizing scientific knowledge and supporting downstream tasks such as literature review generation, knowledge discovery, and trend identification.

\subsubsection*{Acknowledgments}
We thank all the researchers in the community for producing high-quality literature review papers. These articles are the basis of this study.

\bibliographystyle{ACM-Reference-Format}
\bibliography{sample-base}



\appendix
\section{Appendix}\label{sec:appendix}

\subsection{Dataset}\label{app:dataset}

We conduct experiments on 2-hop citation networks for each literature review paper rather than randomly collecting papers to construct a large citation network as a database. This is because using a large, random network complicates performance evaluation, making it difficult to assess both retrieval and clustering accuracy. Additionally, if related papers published after the literature review are present in the citation network, the retriever may include these newer papers in generating the review content. This would lead to an unfair comparison when evaluate the generated content against the human-written review.

\vspace{2pt}
\noindent \textbf{Citation Network Construction Process.}\quad For each literature review, we first extracted its references and constructed a citation tree, with the review paper as the root and its cited papers as the leaves. We then repeated this process for each cited paper, constructing a citation tree for each one. Next, we merged all these trees into a single, large citation network, consolidating any duplicate nodes. To automate this process, we used citation information from arXiv, which provides the LaTeX source code for each paper, including the bib or bbl files. If a paper was available on arXiv, we extracted its .tex file to obtain both the abstract and full text, using these as high-quality text features for the corresponding node. We used the arXiv API to automate this process. For papers not available on arXiv, we used the Google Scholar API to automatically retrieve the abstract, which we used as the text feature for the corresponding node in the citation network. Finally, we removed the node representing the original literature review, leaving 1-hop, 2-hop, and 3-hop citation networks for each review. 
The mutual citations among references form complex citation networks, averaging 6,658.4 papers and 11,632.9 edges, including isolated papers.

\subsection{Implementation}\label{app:imp}

All experiments were conducted on a Linux-based server equipped with 4 NVIDIA A10G GPUs. For 518 review papers, we successfully collected taxonomy trees with hierarchical citation graph clustering labels for 313 of the literature reviews. Of these, 200 reviews were used to train the hierarchical citation graph clustering and taxonomy generation module, while the remaining 118 were used to test the performance of the pre-trained hierarchical citation graph clustering model. 318 reviews were reserved for a comprehensive evaluation of review content generation. The number of articles retrieved in retrieval phase was set to 200. The scaling factor $\alpha$ is set to 1. 

\vspace{2pt}
\noindent \textbf{Pre-Train Hierarchical Citation Graph Clustering Module.}\quad GNN used in this paper is GAT \citep{velivckovic2017graph} which has 2 layers with 4 heads per layer and a hidden dimension size of 1024. MLP$_{\phi}$ has 2 layers and a hidden dimension size of 1024. The edge connection threshold $p_{\tau}$ is searched in [0.1, 0.2, 0.5, 0.8]. 
The clustering model is trained for a maximum of 500 epochs using an early stop scheme with patience of 10. The learning rate is set to 0.001. The training batch is set to 512 and the test batch is 1024. 

\vspace{2pt}
\noindent \textbf{Fine-Tuning.}\quad The LLM backbone is Llama-2-7b-hf. We adopt Low Rank Adaptation (LoRA) \citep{hu2021lora} for fine-tuning, and configure the LoRA parameters as follows: the dimension of the low-rank matrices is set to 8; the scaling factor is 16; the dropout rate is 0.05. For optimization, the AdamW optimizer is used. The initial learning rate is set to 1e-5 and the weight decay is 0.05. Each experiment is run for a maximum of 10 epochs, with a batch size of 4 for both training and testing. The MLP$_{\Phi}$ has 2 layers and a hidden dimension size of 1024.

\vspace{2pt}
\noindent \textbf{LLMs.}\quad When calling the API, we set temperature as 1 and other parameters to default. The content generator is gpt-4o-2024-05-13 and the content judge is and claude-3-haiku-20240307.

\subsection{Evaluation Metrics}\label{app:eval}

We engaged two groups of human raters, each consisting of two professional PhD students in computer science, to evaluate the generated taxonomies and literature reviews using two scoring mechanisms. Each group scores 100 generated samples Given the need to compare model-generated outputs with those written by humans, we designed the evaluation criteria to be straightforward and focused. The two key metrics used are \textbf{Adequacy} and \textbf{Validity}:

\vspace{2pt}
\noindent \textbf{Adequacy} is a binary metric where evaluators respond with \textit{``Yes''} or \textit{``No''} to the question: \textit{``Compared to the taxonomy written by humans, is this taxonomy suitable for learning this field?''} This assesses whether the taxonomy is practically usable and meets the fundamental requirements for understanding the domain.

\vspace{2pt}
\noindent \textbf{Validity}, on the other hand, is rated on a scale from 1 to 5, evaluating the degree to which the taxonomy accurately reflects factual information and represents the domain's conceptual structure. The scoring is defined as follows:
\begin{itemize}[left=0pt]
    \item \textbf{1} -- Completely inaccurate, with significant factual errors or misrepresentations of the domain.
    \item \textbf{2} -- Mostly inaccurate, capturing only a few correct facts but failing to represent the domain coherently.
    \item \textbf{3} -- Moderately accurate, containing some factual correctness but missing important concepts or relationships.
    \item \textbf{4} -- Mostly accurate, representing the domain well with minor factual inaccuracies or omissions.
    \item \textbf{5} -- Highly accurate, thoroughly reflecting the domain's factual structure with no noticeable errors.
\end{itemize}

This combination of metrics allows us to capture both the \textit{practical usability} and the \textit{factual correctness} of the generated taxonomies, ensuring a comprehensive and nuanced evaluation from multiple perspectives.

\begin{figure}[h!]
  \centering
  \includegraphics[width=0.46\textwidth]{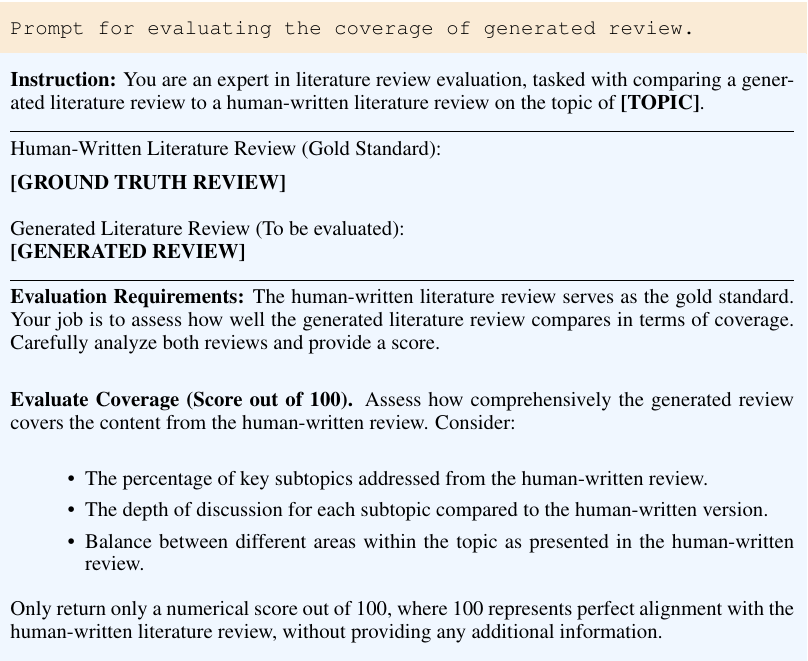}
  \caption{Prompt used for evaluating coverage with LLMs.}
  \label{prompt_converge}
\end{figure}

\begin{figure}[h!]
  \centering
  \includegraphics[width=0.46\textwidth]{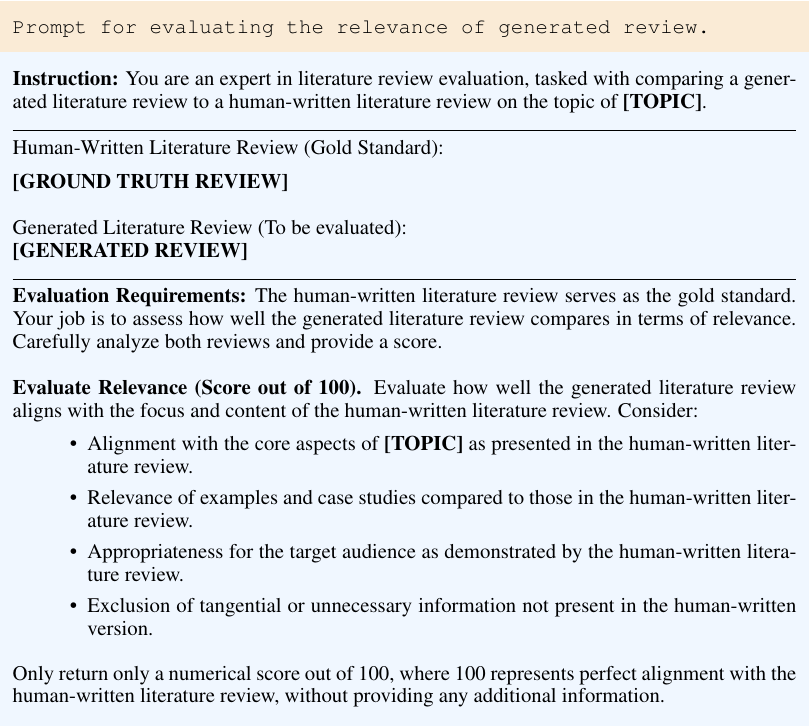}
  \caption{Prompt used for evaluating relevance with LLMs.}
  \label{prompt_relevance}
\end{figure}

\begin{figure}[h!]
  \centering
  \includegraphics[width=0.46\textwidth]{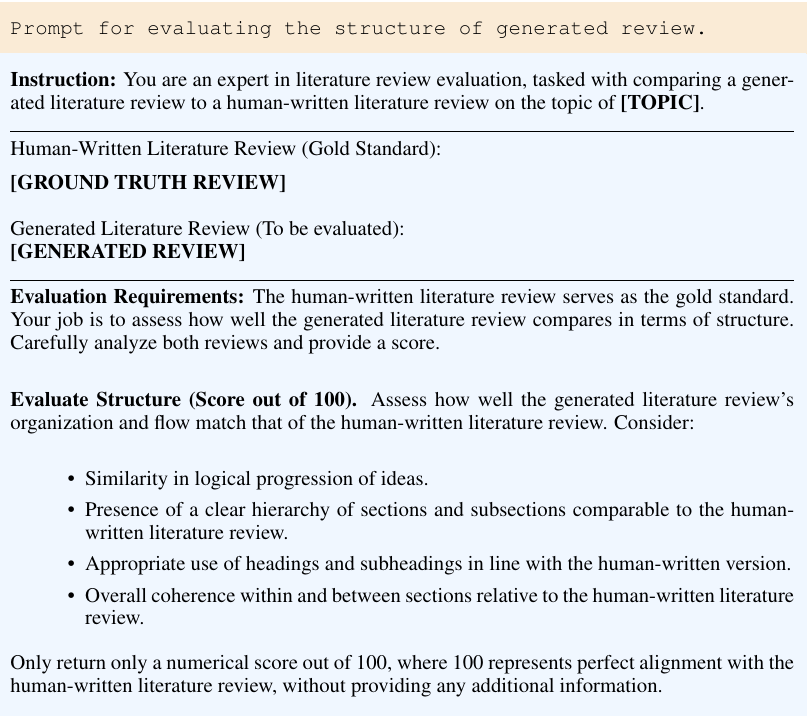}
  \caption{Prompt used for evaluating structure with LLMs.}
  \label{prompt_structure}
\end{figure}

For LLM evaluation, we assess the generated content from three perspectives: coverage, relevance, and structure, with each scored on a scale from 1 to 100. The specific prompts used for these evaluations are shown in \hyperref[prompt_converge]{Figure}~\ref{prompt_converge}, \hyperref[prompt_relevance]{Figure}~\ref{prompt_relevance} and \hyperref[prompt_structure]{Figure}~\ref{prompt_structure}.

\subsection{Investigation of Retrieval Models}\label{app:retrieval}

We experimented with different retrieval models and strategies, testing two representative methods: the sparse retrieval model, BM25 \citep{robertson2009probabilistic}, and the dense retrieval model, SentenceBert \citep{reimers2019sentence}. In citation networks, neighbor information and the topological structure play a crucial role in retrieval, as papers on the same topic often cite each other. To assess the impact of using neighbor information, we applied two retrieval strategies for both models: one incorporating neighbor information as described in \hyperref[sec:retrieval]{Section} \ref{sec:retrieval} (\textit{Retrieval w/ Neighbor}) and the other excluding neighbor information (\textit{Retrieval w/o Neighbor}). Given a topic (specifically, the title of a review paper), we retrieved papers related to this topic from the citation network and measured the accuracy by calculating how many of the retrieved papers appeared in the references of the corresponding literature review. The number of retrieved papers was not fixed, but matched the reference count for each review paper.

\begin{table}[htb]
\centering
\footnotesize
\captionsetup{font=footnotesize, skip=0.10mm}
\caption{Results of retrieval on the citation network corresponding to 50 review papers. \textit{2-hop} and \textit{3-hop} represent citation networks of review papers at different scales. \textit{1-hop (merged)} refers to the 1-hop citation network of a review paper, merged with all other 1-hop citation networks, different review papers. Similarly, \textit{2-hop (merged)} is constructed by merging the 2-hop citation network of a review with all other 49 review papers.}\label{tab:retrieval}
\begin{tabular}{l|cccc}
\hline
\textbf{\multirow{2}*{Model}} &  \multicolumn{4}{c}{{\fontfamily{pcr}\selectfont Accuracy$\uparrow$}} \\ 
 \cline{2-5} & 1-hop (merged) & 2-hop & 2-hop (merged)  & 3-hop \\ \hline
\multicolumn{5}{c}{{\fontfamily{pcr}\selectfont Retrieval w/o Neighbor}} \\ \hline
\textbf{BM25} & 0.3308 & 0.1375 & 0.0947  &  0.1014 \\ \hdashline
\textbf{SentenceBert}  & 0.5234  & 0.1746 & 0.1521 & 0.1490 \\ \hline
\multicolumn{5}{c}{{\fontfamily{pcr}\selectfont Retrieval w Neighbor}} \\ \hline
\textbf{BM25} & 0.7445 & 0.6435 & 0.5950 &  0.6179 \\ \hdashline
\textbf{SentenceBert} & 0.2602 & 0.2758 & 0.2181 & 0.2144 \\
\hline
\end{tabular}
\end{table}

As shown in \hyperref[tab:retrieval]{Table} \ref{tab:retrieval}, SentenceBert consistently outperforms BM25 across all scales when neighbor information is not used. For example, in the \textit{1-hop merged} case, SentenceBert achieves an accuracy of 0.5234, significantly higher than BM25's 0.3308. However, both methods show relatively low accuracy without neighbor information, and their performance declines as the size of citation networks increases, indicating that retrieving relevant papers becomes more challenging as the network expands. In contrast, BM25 significantly outperforms SentenceBert when neighbor information is utilized. For instance, in the \textit{1-hop merged} case, BM25 reaches an accuracy of 0.7445, while SentenceBert's accuracy drops sharply to 0.2602. BM25 maintains much higher accuracy across all scales with neighbor information. BM25, as a sparse retrieval model, relies on exact term matches, which is particularly advantageous in structured environments like citation networks, where specific terms (e.g., paper titles or keywords) are highly relevant. The inclusion of neighbor information allows BM25 to better capture relationships between papers by focusing on direct term matches in titles or citations. When neighbor information is introduced, the context around the target paper becomes more critical. BM25 effectively leverages this by prioritizing exact matches from neighboring papers, while SentenceBert, which focuses on semantic similarity, may lose precision when handling a broader context that includes less directly related papers.

Without graph-aware retrieval, methods like AutoSurvey must retrieve a large number of papers (e.g., 1200 in AutoSurvey) to avoid missing relevant ones. Retrieving fewer papers risks missing important content, while retrieving too many introduces noise from irrelevant papers. Graph-aware retrieval significantly alleviates this issue. The graph context-aware retrieval strategy we propose achieves more accurate results with fewer retrievals, i.e., 200, reducing irrelevant information and contributing to the superior generation performance of our model. Moreover, even when applied to large citation networks (such as \textit{2-hop merged} each containing over 200,000 papers), our method maintains stable retrieval accuracy, demonstrating HiReview's robustness across different citation network sizes. Additionally, we experimented with different retrieval strategies, such as retrieving based on both the title and abstract. We found that using only the title yielded the best results, while incorporating additional information like the abstract reduced retrieval performance.

\subsection{The Choice of GNN}\label{app:gnn}

In addition to GAT \citep{velivckovic2017graph}, we also explored other GNNs as graph encoders, i.e., GCN \citep{kipf2016semi} and Graph Transformer \citep{shi2020masked}. The comparison results of these models on clustering are shown in \hyperref[tab:gnn]{Table} \ref{tab:gnn}.

\begin{table}[htb]
\centering
\footnotesize
\captionsetup{font=footnotesize, skip=0.10mm}
\caption{Performance of different GNN on hierarchical citation graph clustering.}
        \begin{tabular}{l|ccc}
        \hline
        \textbf{Model} &  Level 1 & Level 2 & Average \\\hline
        \textbf{GAT} & 0.7127 & 0.6395 &  0.6761 \\ \hdashline
        \textbf{GCN} & 0.6730 & 0.5963  & 0.6347 \\ \hdashline
        \textbf{Graph Transformer} & 0.6811 & 0.6024  & 0.6418 \\ 
        \hline
        \end{tabular}
\label{tab:gnn}
\end{table}

GAT achieves the highest performance across both levels, with an average score of 0.6761. It outperforms the other models at both Level 1 (0.7127) and Level 2 (0.6395), making it the most effective GNN for this task. This superior performance can likely be attributed to GAT's attention mechanism, which enables the model to assign varying importance weights to neighboring papers, allowing it to better capture the hierarchical structure of the graph. As a result, we selected GAT as the GNN backbone for HiReview.

\section{Literature Review Generation Example}\label{app:example}

\begin{figure*}[t!]
  \centering
  \includegraphics[width=\textwidth]{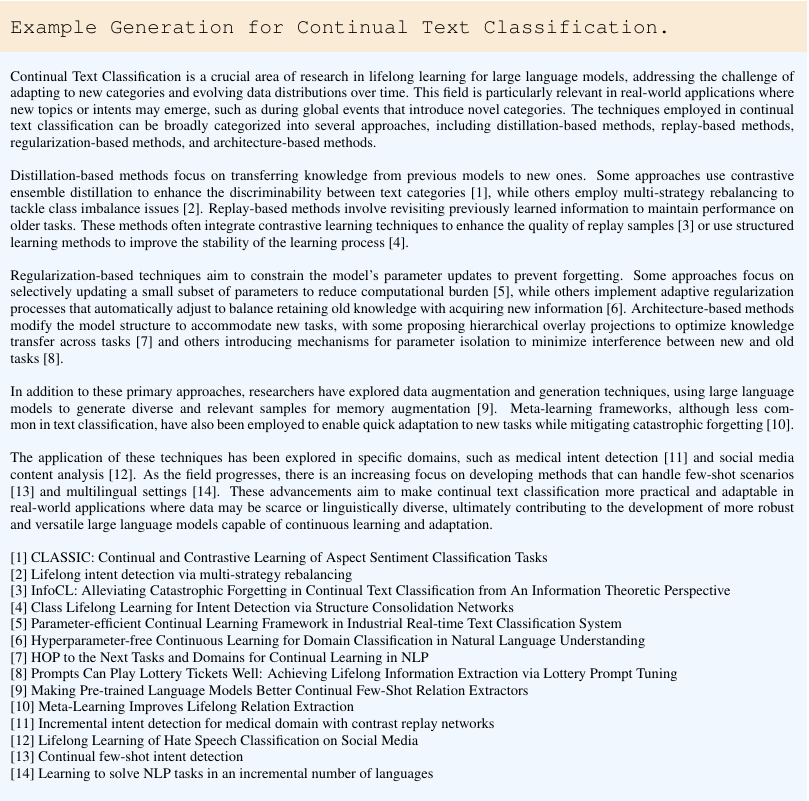}
  \caption{Example of a generated section on the topic of Continual Text Classification.}
  \label{fig:gen}
\end{figure*}

\end{document}